\documentclass{article}

\usepackage[preprint]{neurips_2023}

\usepackage[utf8]{inputenc} 
\usepackage[T1]{fontenc}    
\usepackage{hyperref}       
\usepackage{url}            
\usepackage{booktabs}       
\usepackage{amsfonts}       
\usepackage{nicefrac}       
\usepackage{microtype}      
\usepackage[dvipsnames]{xcolor}
\usepackage{graphicx}
\usepackage{subcaption}

\usepackage{amsmath}
\usepackage{amssymb}
\usepackage{mathtools}
\usepackage{amsthm}
\theoremstyle{plain}

\theoremstyle{definition}

\theoremstyle{remark}

\usepackage[capitalize,noabbrev]{cleveref}

\newcommand{\defeq}{\mathrel{\mathop:}=}

\newcommand{\R}{\mathbb{R}}

\newcommand{\red}{\color{red}}
\newcommand{\green}{\color{ForestGreen}}
\newcommand{\orange}{\color{Orange}}

\title{Auditing Visualizations: Transparency Methods Struggle to Detect Anomalous Behavior}

\author{%
	Jean-Stanislas Denain \& Jacob Steinhardt \\
  UC Berkeley\\
}

\begin{document}
\maketitle

\begin{abstract}
Model visualizations provide information that outputs alone might miss. 
But can we trust that model visualizations reflect model behavior? 
For instance, can they diagnose abnormal behavior such as planted backdoors or overregularization? 
To evaluate visualization methods, we test whether they assign different visualizations to anomalously trained models and normal models.
We find that while existing methods can detect models with starkly anomalous behavior,
they struggle to identify more subtle anomalies. 
Moreover, they often fail to recognize the inputs that induce anomalous behavior, e.g.~images containing a spurious cue. 
These results reveal blind spots and limitations of some popular model visualizations. 
By introducing a novel evaluation framework for visualizations, our work paves the way for developing more reliable model transparency methods in the future\footnote{Our code for the experiments is released at \url{https://github.com/js-d/auditing-vis}.}.
\end{abstract}

\section{Introduction}
\label{introduction}
Neural networks with similar validation accuracy can behave very differently when deployed. A network can have a high validation accuracy, but be vulnerable to distribution shift, backdoors, or spurious cues learned during training. 
To diagnose these failures, model visualizations rely on the internals of a network, such as the activations of inner layers or the model's gradients \citep{molnar2022}. By opening the black box, they may better uncover failure modes \citep{milan2022}, 
tell us where to intervene to repair models \citep{wong2021leveraging, santurkar2021editing}, or provide auxiliary training objectives \citep{conceptbottleneck}. 

However, model visualizations can only fulfill these promises if they actually track model behavior. 
Previous evaluations have shown that many visualization methods fail basic tests of validity \citep{sanity18, adebayo2020debugging}.
However, we seek more than simply pointing out failures in methods: we would like a quantifiable goal for interpretability methods to aspire to.

Following \citet{adebayo2020debugging}, we focus on the goal of debugging model errors: we say that a method is good at debugging if it can \textbf{successfully uncover a wide range of possible anomalous behaviors} in models.
To measure this, we train a wide variety of models with anomalous behaviors, such as models containing backdoors or trained with spurious cues. 
We also train a set of reference models, that do not exhibit anomalous behavior, and we test whether visualization methods behave differently on the reference and anomalous models.

\begin{figure*}[t]
    \centering
    \includegraphics[width=350px]{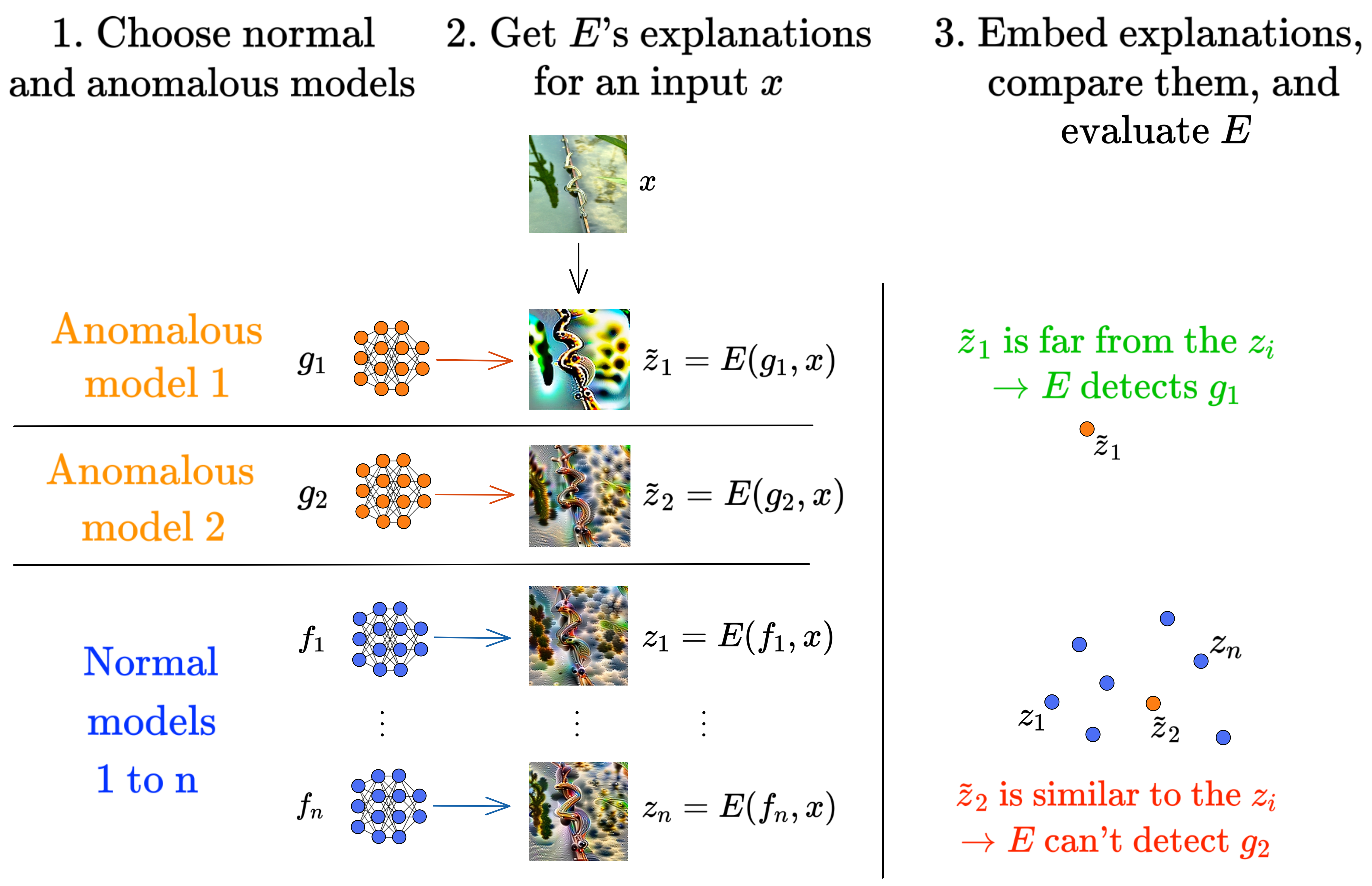}    
    \caption{Testing whether a method \(E\) can detect if a model is anomalous (\cref{detection}). Here \(E\) can detect that \(g_1\) is anomalous, but not \(g_2\).}
    \label{detection_task_figure}
\end{figure*}

Our pipeline for evaluating visualizations is illustrated in \cref{detection_task_figure}. 
We leverage the fact that visualizations are images and thus can be embedded in a semantic feature space. 
Under our evaluation metric, a visualization method distinguishes anomalous models if its 
embedding for the anomalous model is far from the reference models, relative to the variation within the reference models. 
We use this metric to measure whether visualizations can (1) \emph{detect} that a network is anomalous, and 
(2) determine how the model is unusual, i.e.~\emph{localize} the inputs where it behaves atypically.

We evaluate five widely used model visualization techniques: Caricatures \citep{olah_caricatures}, Integrated Gradients \citep{sundararajan2017axiomatic}, Guided Backpropagation \citep{springenberg2015striving}, Grad-CAM, and Guided Grad-CAM \citep{Selvaraju_2019}. 
For each technique, we test whether it can detect 24 anomalous models across 7 categories: adversarial training \citep{madry2019deep}, randomized smoothing \citep{cohen2019certified}, shape bias \citep{geirhos2019imagenettrained}, backdoors \citep{li2021backdoor}, spurious features, training without data from certain classes, and training on 
a face-obfuscated version of ImageNet \citep{yang2021study}.

Our main takeaway is that model visualizations fail to detect and localize a large number of anomalies.
Most methods only consistently detect anomalous models in the regime where the validation accuracy itself is lower than normal (\emph{stark} anomalies, such as adversarial training), and struggle with anomalies that affect behavior on a smaller number of inputs (\emph{subtle} anomalies, for example removing all the images in a single class from the training data). 
Moreover, methods often fail to localize the inputs for which behavior is different, such as images containing a spurious cue for the spurious features anomaly.

Taken together, our results underscore the need to systematically evaluate visualization methods to check that they track model behavior. 
If we view visualization methods as a way of auditing models, our detection and localization benchmarks are a form of 
\emph{counterauditing}---planting intentionally corrupted models to check that they are discovered. 
To achieve high reliability in machine learning, such counteraudits are a crucial counterpart to regular audits.
Our framework can easily be adapted to other visualization methods and anomalies, and we encourage the community to use such evaluations in the future.

\section{Related Work}
\label{related}
Our work is closely related to \citet{sanity18} and \citet{adebayo2020debugging}, which evaluate feature attribution methods by their ability to diagnose different kinds of model errors. 
Similarly, \citet{yilunzhou} and \citet{bastings} study whether feature attributions detect spurious features, and \citet{adebayo2022post} shows that model explanations can't uncover unknown spurious correlations.
Compared to these works, our evaluation framework is more flexible: given a set of reference models and a distance metric between model visualizations, it can work for any visualization method, model anomaly, and any modality (vision, language, or other).
This flexibility allows us to consider a much greater variety of model anomalies (24 anomalous models), which provides more detailed information about the limitations of visualization methods. 
It also makes it possible to compare feature attributions and different kinds of methods, such as feature visualizations.

\citet{lin2020see} study whether 7 feature attribution methods, including Guided Backpropagation, are able to localize backdoor triggers: unlike us, they find that most methods fail, however they consider more sophisticated trojan attacks.
Following our work, \citet{casper2023benchmarking} also used trojan discovery to benchmark feature attributions and feature synthesis methods, and found that most feature attribution methods could approximately locate the trigger, although failed to beat a simple baseline (see Figure 2 in \citep{casper2023benchmarking}.
In contrast with these works, we consider anomalies other than backdoors, we define localization in a different and more general way, and we also assess whether a method can detect that a model is anomalous.

\begin{figure*}[t]
	\begin{center}
		\includegraphics[width=400px]{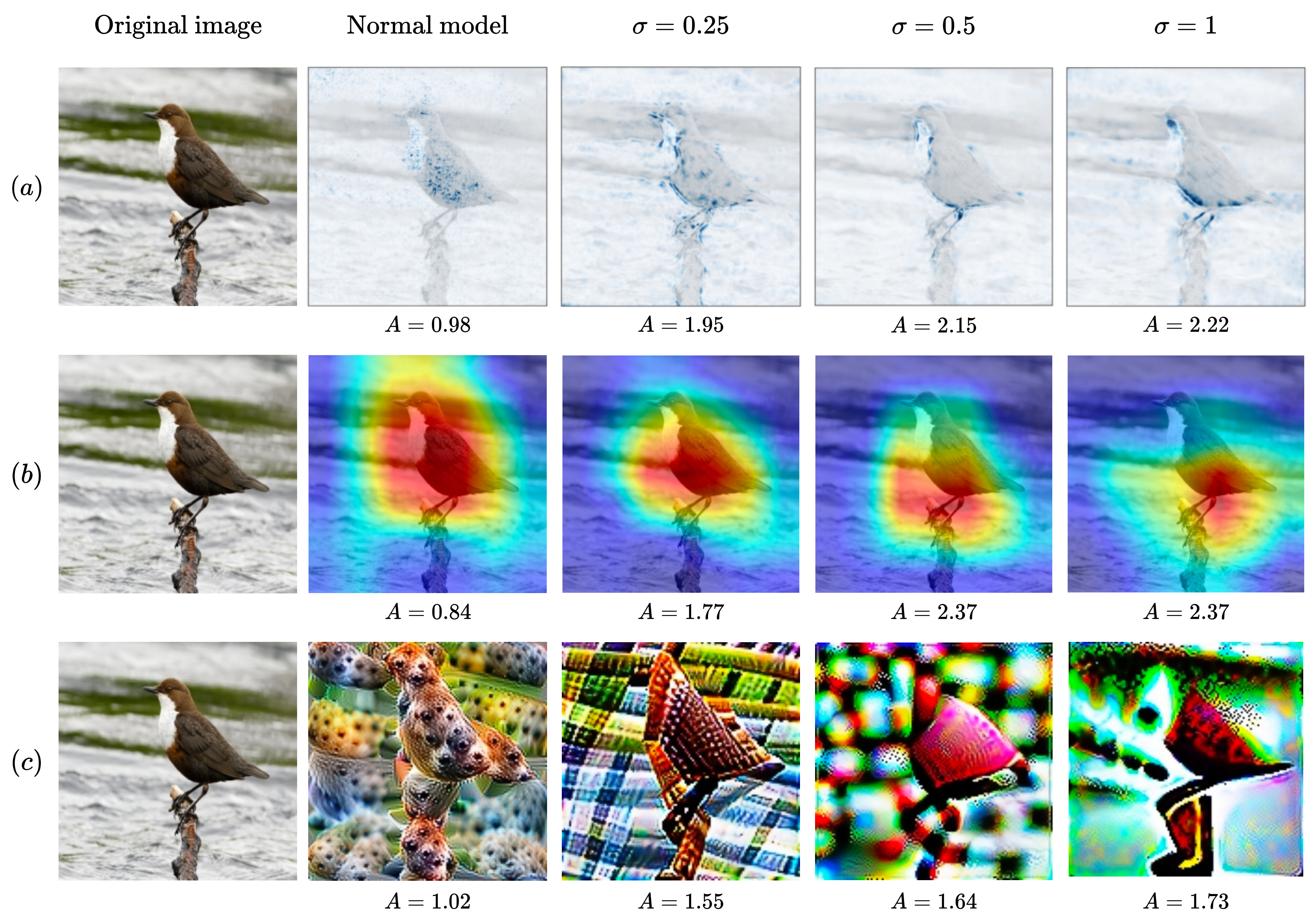}    
		\caption{All methods are able to detect stark anomalies such as randomized smoothing (adding Gaussian noise to training inputs). We show model explanations by (a) Integrated Gradients, (b) Grad-CAM, and (c) Caricatures (layer (3,2,2)). We consider a normal model and anomalous models trained with increasing amounts $\sigma$ of randomized smoothing. The explanations of the anomalous models appear very different from those of the normal model, and their anomaly scores $A$ are greater.}
		\label{detect_success}
	\end{center}
	\vskip -2em
\end{figure*}

\citet{ding2021grounding} apply a similar framework to evaluate representation similarity metrics: just like the model visualizations of an anomalous model should be different from those of a normal model, a representation similarity metric should assign a high distance between representations with different functional properties.

Finally, there are other approaches to ground and evaluate transparency methods. One especially ambitious goal is to reverse engineer entire parts of neural networks: a preliminary example is \citet{cammarata2021curve}, in which the authors use feature visualization methods to hand-code a curve detection module in InceptionV1. A related perspective is model repair: for example, \citet{wong2021leveraging} and \citet{singla2021salient} both introduce new visualization-based methods to help humans identify and repair spurious features in models. Finally, another way of evaluating transparency methods is to run human subject studies to determine how much they help humans understand the underlying models \cite{doshivelez2017rigorous,bhatt2020explainable}.

\section{Problem Setup}
\label{setup}
\vskip -0.5em
In this section, we first introduce the anomalous models and reference models used in our evaluations (\cref{anomalous-models}), then define model visualizations and visualization anomaly scores (\cref{anomalous-visualizations}). Finally, we introduce the transparency methods considered in this work (\cref{visualization-methods}).

\subsection{Anomalous models and reference models} 
\label{anomalous-models}

We call a model \(f\) anomalous if it differs from a set of reference models that specify normal behavior, for example in its accuracy on an out-of-distribution test set.

\textbf{Reference models.} In what follows, all the models we consider will be ResNet-50 variants, and our reference models will be \(n=12\) ResNet-50 networks \(g_1, \dots, g_n\) trained on ImageNet with the same hyperparameters but different random seeds.

\textbf{Anomalous models.}
For the choice of the anomalous model \(f\), we consider 24 ResNet-50 models comprising 7 different types of anomalies.
The first three families of anomalies are \emph{stark} (they always induce a significant drop in validation accuracy), while the 
remainder are \emph{subtle} (their behavior is only different on out-of-distribution data).
We will often find that model visualizations can detect stark anomalies, but not the subtle ones. We describe each family below, with full training details given in \cref{more-details}.

\emph{Randomized Smoothing:} These models are trained on ImageNet images perturbed by isotropic Gaussian noise with standard deviation $\sigma$, for $\sigma \in \{0.25, 0.5,1.0\}$. We obtained checkpoints from the ImageNet Testbed \citep{taori2020measuring}. Randomized smoothing substantially decreases the test set accuracy, to at most $0.4$ for \(\sigma=0.25\). Unlike in applications to adversarial robustness, we do not add noise to the inputs at evaluation time.

\emph{Adversarial Training:} These are robust models adversarially trained on ImageNet with a PGD adversary. We obtained models with \(\ell^\infty\)-norm constraints of \(\epsilon=4\) and \(8\) from \citet{robustness}, and models with \(\ell^2\)-norm constraints of \(\epsilon=0.01\), \(0.1\), \(1\), and \(5\) from \citet{salman2020adversarially}. Most of these models exhibit a noticeable drop in validation accuracy. 

\emph{Shape Bias:} This anomalous model is trained on Stylized-ImageNet \citep{geirhos2019imagenettrained}, which makes it more sensitive to shapes and less sensitive to textures when classifying images. Once again, the resulting model has low accuracy on the ImageNet validation set ($0.602$).

\emph{Spurious Features:} We trained models on ImageNet, but where a fraction \(\delta\) of each training batch's images had the class index overlaid at the bottom right of the image. We trained 4 such models with \(\delta \in \{1.0, 0.5, 0.3, 0.1\}\).
The spurious features have a large impact on validation accuracy when \(\delta=1.0\), but the effect nearly disappears for smaller $\delta$.

\emph{Backdoors:} We trained 6 backdoored ResNets with a pixel-modification data poisoning attack by following the methodology in \citet{bagdasaryan2020blind}. These models exhibit a negligible drop in validation accuracy (at most $0.01$ and sometimes smaller). On the other 
hand, their out-of-distribution accuracy on images with the backdoor trigger is near zero.

\emph{Removing data from a class:} We trained networks on variants of ImageNet with all images from a given class removed. We considered 3 such models, removing data from classes 414 (\texttt{backpack}), 400 (\texttt{academic gown}), and 218 (\texttt{Welsh springer spaniel}). We sampled classes 400 and 414 uniformly from the 1000 classes. Because ImageNet contains so many dogs, we thought that removing data from a dog class would be harder to detect, so we chose class 218 by sampling uniformly from dog classes. These anomalies have a negligible effect on the overall validation accuracy, but drop the accuracy on the missing class to 0.

\emph{Training on blurred images:} Finally, we trained a model on a variant of ImageNet with blurred faces \citep{yang2021study}. This decreases the validation accuracy by less than $1\%$, but has a large effect on some classes, such as \texttt{harmonica}, where key 
information becomes blurred.

\begin{table*}
	\begin{center}
    \caption{Fraction of anomalous models detected by each method. All methods succeed at the detection task for the stark anomalies, but fail for most of the subtle anomalies.}
    \label{tab:short_det_alex}
\begin{tabular}{lllllllll}
    \toprule
    & \multicolumn{4}{c}{Stark Anomalies} & \multicolumn{4}{c}{Subtle Anomalies} \\ \cmidrule(lr){2-5} \cmidrule(lr){6-9}
Anomaly Type & Smoothing  & Shape Bias & Adv. ($\ell^\infty$) & Adv. ($\ell^2$) & Backdoor  & Spurious &  Blur & Missing \\
\midrule
GBP                & \green{\textbf{1}} & \green{\textbf{1}} & \green{\textbf{1}} &  \orange{\textbf{0.75}} &     \orange{\textbf{0.67}} & \orange{\textbf{0.5}} &  \red{\red{\textbf{0}}} &    \red{\textbf{0}}             \\
IG            & \green{\textbf{1}} & \green{\textbf{1}} & \green{\textbf{1}} &  \green{\textbf{1}} &     \orange{\textbf{0.17}} & \green{\textbf{1}} &  \red{\textbf{0}} &    \red{\textbf{0}}             \\
GC            & \green{\textbf{1}} & \green{\textbf{1}} & \green{\textbf{1}} &  \orange{\textbf{0.75}} &     \orange{\textbf{0.5}} & \orange{\textbf{0.25}} &  \red{\textbf{0}} &    \red{\textbf{0}}             \\
GGC     & \green{\textbf{1}} & \green{\textbf{1}} & \green{\textbf{1}} &  \orange{\textbf{0.75}} &     \orange{\textbf{0.5}} & \orange{\textbf{0.25}} &  \red{\textbf{0}} &    \red{\textbf{0}}             \\
Caricatures        & \green{\textbf{1}} & \green{\textbf{1}} & \green{\textbf{1}} &  \green{\textbf{1}} &     \green{\textbf{1}} & \orange{\textbf{0.25}} &  \red{\textbf{0}} &    \red{\textbf{0}}             \\
\bottomrule
\end{tabular}
\end{center}
\vskip -1.5em
\end{table*}

\subsection{Anomaly scores for visualizations} 
\label{anomalous-visualizations}

Recall that model visualizations \(E(\cdot ,\cdot)\) are functions that take as arguments a neural network $f$ and input $x$, 
and output some information $E(f, x)$ about how $f$ processes $x$. This information is usually a vector or tensor, e.g.~an image 
in the case of feature visualizations.

We will call an visualization \(z = E(f, x)\) anomalous when it is far away from reference visualizations \(z_i = E(g_i, x)\). 
Given a metric \(\mathrm{d}(\cdot,\cdot)\) between visualizations, a simple way to quantify this is via the average distance between \(z\) and the reference visualizations, normalized by the average distance between the reference visualizations themselves:
\begin{equation}
A(f, E, x) \defeq \frac {\frac{1}{n} \sum\limits_{i=1}^n \mathrm{d}(z, z_i)} {\frac{2}{n(n-1)} \sum\limits_{1 \leq i < j \leq n} \mathrm{d}(z_i, z_j)}.
\end{equation}
We call this ratio the \textbf{anomaly score} of \(z\) compared to the reference visualizations \(z_1, \dots, z_n\).\footnote{We omit the dependence on $\mathrm{d}$ and $g_{1:n}$ in the notation $A(f, E, x)$ because we only consider one metric outside of the Appendix, and the reference models are fixed throughout this paper.} When this ratio is much larger than 1, then the visualization \(z = E(f, x)\) is anomalous. A good transparency method should assign a high anomaly score to models with anomalous behavior.

For the metric \(\mathrm{d}(\cdot, \cdot)\), we use the LPIPS distance, since \citet{zhang2018unreasonable} show that it aligns well with human perceptual similarity. In the main text, we used AlexNet features for the LPIPS distance, because \citet{zhang2018unreasonable} found that they perform the best. As a robustness check, we also computed anomaly scores for LPIPS distances with two other networks, VGG and SqueezeNet; we show the corresponding results in the Appendix.

\subsection{Visualization methods} \label{visualization-methods}

We assess four feature attribution methods and one feature visualization method.


\textbf{Feature attributions.} Consider a neural network classifier $f: \R^d \to \R^C$, which outputs a logit for each possible class $1 \leq c \leq C$. For an input $x$, a feature attribution of $f$ with target class \(c\) is a tensor with the same shape as $x$. Intuitively, it represents the relevance of each of \(x\)'s entries to \(f(x)\)'s output logit for class $c$.
In what follows, we'll always take $c$ to be the output class $c_\text{max} = \underset{c}{\text{argmax}} f_c(x)$. We consider four feature attribution methods:

\emph{Guided Backpropagation} (GBP) \citep{springenberg2015striving} takes the gradient \(\nabla_x f_c(x)\) of the output logit with respect to the pixels, and modifies it by only backpropagating non-negative gradients through ReLU nonlinearities.

\emph{Integrated Gradient} (IG) \citep{sundararajan2017axiomatic} attributions take the integral of the gradient of \(f_c\) along a linear path connecting \(x\) to some base point \(x_0\):   
    \begin{equation}
    \left(x-x_0\right) \times \int_{0}^{1} \nabla_x f_c\left(x_0+\alpha\left(x_{i}-x_0\right)\right) d \alpha.
    \end{equation}

\emph{Grad-CAM} (GC) \citep{Selvaraju_2019} highlights large regions of the input image that are important for the prediction. It is the sum of the activations of the final convolutional layer's feature maps, weighted by the gradient of \(f_c\) with respect to each feature map.

\emph{Guided Grad-CAM} (GGC) \citep{Selvaraju_2019} takes the product of guided backpropagation attributions and GradCAM attributions.

We obtained Guided Backpropagation, Integrated Gradients and Guided Grad-CAM attributions using the Captum library \citep{kokhlikyan2020captum} with default settings. We obtained Grad-CAM attributions using the \texttt{pytorch-grad-cam} library \citep{jacobgilpytorchcam}. We display these visualizations by superposing the attributions on top of the input image (c.f.~the first two rows of \cref{detect_success}).
 
\begin{figure*}[t]
    \centering
    \includegraphics[width=400px]{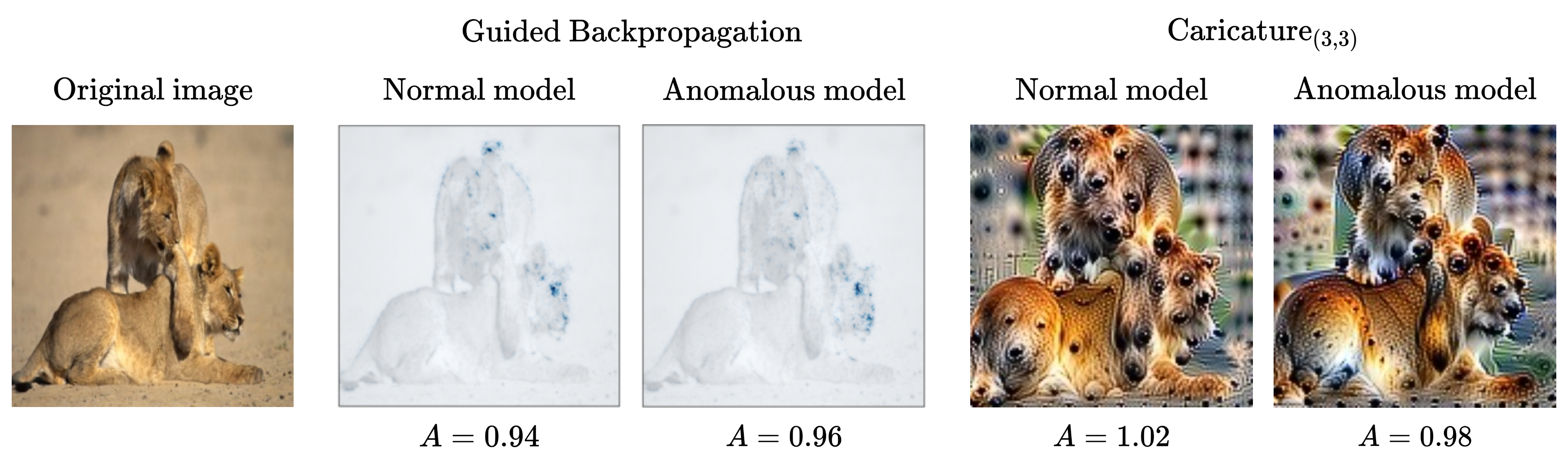}    
    \caption{Most methods struggle to detect subtle anomalies like Spurious Features. We show model explanations by Guided Backpropagation and Caricatures (layer (3,3)) for a normal model and a Spurious Features model ($\delta = 0.3$), with each explanation's anomaly score $A$. Both methods fail to detect that the model is anomalous, and the explanations of both models appear very similar.}
    \label{detect_fail}	
	\vskip -1em
\end{figure*}


\textbf{Feature Visualization: Caricatures.} For a neural net classifier \(f\), let \(f_\ell, 1 \leq \ell \leq L\) denote its layers, i.e.~$f_\ell$ maps an input $x$ to the vector of activations at layer $\ell$.
Intuitively, a Caricature
describes what a layer ``sees'' in an input \(x\) by visualizing the activation vector \(f_\ell(x)\).
 
Formally, the \emph{Caricature} of an input $x$ at layer $\ell$ is the solution to the following optimization problem \citep{olah_caricatures}: 
$\text{Caricature}_{\ell}(x, f) := \text{argmax}_{x'} \langle f_\ell(x'), f_\ell(x) \rangle$.
In practice, \(\text{Caricature}_{\ell}(x, f)\) solves a regularized, preconditioned, and data-augmented version of this optimization problem \citep[see][``The Enemy of Feature Visualization'']{olah2017feature}. We obtained Caricatures (c.f.~\cref{detect_success}c) by running 512 steps of optimization with the Lucent \citep{lucent} library's default parameters.

\section{Detection}
\label{detection}
In this section, we formally define our detection metric, then describe our experimental procedure. Finally, we present and interpret 
our empirical results.

\textbf{Detection Metric.} To test whether a method $E(\cdot, \cdot)$ can detect an anomalous model, we compute visualizations on a set of $m$ sample images $\{x_1, \ldots, x_m\}$. 
We define the overall anomaly score of a model, $\bar{A}(f, E)$, as the average of the anomaly scores for the visualizations 
$E(f, x_1), \ldots, E(f, x_m)$. 
We say that \(E\) \textbf{detects} \(f\)
if $\bar{A}(f, E)$ is larger than all the normal models' anomaly scores $\bar{A}(g_i, E)$.

\textbf{Experimental Procedure.}
We sampled \(m=50\) input images uniformly from the ImageNet test set as our sample images \(x_j\).
To reduce computation time for Caricatures, we visualized a subset of $N_\ell=7$ layers in different parts of the model: 
layers (2,2,1), (3,0,3), (3,2,2), (3,3), (3,4,3), (4,0,3), and (4,3,2).\footnote{Here layer $(i,j,k)$ is convolutional layer $k$ in block $j$ of stage $i$. Unlike the other layers, layer (3,3) comes after, rather than within, a residual block.} We chose these layers to have information about different parts of the network: in earlier or later stages, in earlier or later residual blocks within a stage, and after or within residual blocks. 
For concision, we averaged the anomaly scores over these 7 layers as well as the 50 images to obtain an overall anomaly score for all caricatures \(\bar{A}(f, \text{Caricatures})\). We show individual layer results in the Appendix. Integrated Gradients, Guided Backpropagation, Grad-CAM and Guided Grad-CAM do not depend on a layer: for them, we just average over images and report \(\bar{A}(f, \text{IntGrad})\) and \(\bar{A}(f, \text{GBP})\).

\textbf{Results for stark anomalies.}
For every transparency method \(E\) and every family of anomalous models, we report the fraction of anomalous models that \(E\) detects in \cref{tab:short_det_alex}. The five methods are able to detect that all the models trained on Stylized ImageNet or with randomized smoothing are anomalous. They also usually detect that adversarially trained models are anomalous, although GBP, Grad-CAM, and Guided Grad-CAM fail to detect $\ell^2$-adversarial training with radius \(\epsilon = 0.01\). 

Qualitatively inspecting the visualizations corroborates these results. \cref{detect_success} shows the Integrated Gradients, Grad-CAM, and Caricature$_{(3,2,2)}$ visualizations of the randomized smoothing models and a normal model. The visualizations of the anomalous models appear very different from those of the normal model. Moreover, they appear more different from the normal visualizations as \(\sigma\) increases.

\textbf{Results for subtle anomalies.} The results are much worse for the subtle anomalies. No methods detect that any of the networks trained on blurred faces or with missing data are anomalous. For backdoors and spurious cues, the detection results differ between methods: for example, Caricatures detect that all backdoored models are anomalous but only a quarter of models in the Spurious Features category, whereas Integrated Gradients detect that all the models that rely on spurious features are anomalous but only 17$\%$ of backdoored models. 

\begin{table*}
	\begin{center}
    \caption{Localization AUROC (\%) for every anomaly and explanation method, with AUROCs higher than 70\% bolded. Although feature attributions successfully localize the backdoors, no method can localize all of the anomalies, and visualizations often have close to random performance, with $54\%$ of AUROCs between 40\% and 60\%.}
    \label{tab:short_loc_alex}
\begin{tabular}{lllllllllllllll}
    \toprule
    Category & \multicolumn{6}{c}{Backdoor} & \multicolumn{3}{c}{Missing class} & \multicolumn{4}{c}{Spurious} & Blur   \\
    \cmidrule(lr){2-7} \cmidrule(lr){8-10} \cmidrule(lr){11-14}
Anomaly &   0 &   1 &   2 &   3 &   4 &   5 & 218 & 400 & 414 & 0.1 & 0.3 & 0.5 & 1.0 & \ \\
\midrule
GBP                &  \textbf{100} &   \textbf{100} &   \textbf{100} &  \textbf{100} &   \textbf{100} &   \textbf{100} &  69 &  43 &  58 &  \textbf{72} &  \textbf{77} &  \textbf{80} &  69 & \textbf{87}\\
IG            &  \textbf{99} &    \textbf{97} &    \textbf{98} &   \textbf{98} &    \textbf{99} &    \textbf{94} &  57 &  61 &  46 &  \textbf{82} &  \textbf{77} &  \textbf{85} &  42 & \textbf{85}\\
GC            &  \textbf{98} &    \textbf{99} &    \textbf{95} &   \textbf{98} &    \textbf{99} &    \textbf{97} &  \textbf{81} &  58 &  59 &  53 &  55 &  53 &  53 & \textbf{76}\\
GGC     &  \textbf{99} &   \textbf{100} &    \textbf{96} &  \textbf{100} &   \textbf{100} &   \textbf{100} &  \textbf{74} &  40 &  58 &  51 &  70 &  64 &  57 & \textbf{79}\\
{Car}$_{(2,2,1)}$ &  37 &    40 &    50 &   45 &    52 &    50 &  38 &  45 &  67 &  57 &  48 &  57 &  52 &  28 \\
{Car}$_{(3,0,3)}$ &  47 &    54 &    54 &   53 &    50 &    60 &  62 &  52 &  51 &  49 &  54 &  40 &  51 &  65 \\
{Car}$_{(3,2,2)}$ &  56 &    53 &    56 &   52 &    55 &    50 &  60 &  61 &  41 &  50 &  52 &  51 &  49 &  66 \\
{Car}$_{(3,3)}$       &  51 &    56 &    59 &   60 &    \textbf{71} &    68 &  56 &  47 &  62 &  57 &  55 &  46 &  53 &  \textbf{82}\\
{Car}$_{(3,4,3)}$ &  57 &    59 &    52 &   58 &    \textbf{71} &    63 &  53 &  58 &  57 &  42 &  43 &  46 &  39 &  53 \\
{Car}$_{(4,0,3)}$ &  64 &    \textbf{80} &    67 &   48 &    \textbf{77} &    \textbf{72} &  66 &  32 &  46 &  52 &  54 &  46 &  \textbf{77} &  59 \\
{Car}$_{(4,2,3)}$ &  \textbf{84} &    \textbf{90} &    \textbf{91} &   \textbf{93} &    \textbf{82} &    69 &  51 &  46 &  50 &  55 &  51 &  52 &  64 &  \textbf{73} \\
\bottomrule
\end{tabular}
\end{center}
\vskip -1em
\end{table*}

Once again, qualitatively inspecting the visualizations confirms these findings. For example, \cref{detect_fail} shows visualizations by Guided Backpropagation and Caricature$_{(3,3)}$ of a normal model and a Spurious Features model with \(\delta = 0.3\). Both methods fail to detect that this model is anomalous, and for both methods the visualizations of the two models appear very similar.

In summary, most methods succeed at detecting stark anomalies, but fail for most of the subtle anomalies. Grad-CAM and Guided Grad-CAM perform the same, and generally worse than Integrated Gradients, Guided Backpropagation, and Caricatures. We show in the Appendix that these results are robust with respect to the choice of the metric between visualizations (\cref{appendix_detection}).

\section{Localization}
\label{localization}
Having studied transparency methods' ability to detect that a model is anomalous, we now turn to the question of localization.

\textbf{Localization Metric.} Given an anomalous model \(f\), we assess a method \(E(\cdot, \cdot)\)'s localization ability via the question: is the anomaly score $A(f, E, x)$ highest on the inputs $x$ where $f$'s behaviour is unusual? We will measure \(E\)'s ability to distinguish these inputs via AUROC. Concretely, we sample \(p=40\) ``normal'' inputs (e.g.~from the ImageNet validation set), and \(p\) ``abnormal'' images on which the anomalous 
network has abnormal behavior (e.g.~images with a backdoor trigger). We then compute anomaly scores for all 
80 images.

Thresholding on the anomaly score defines a binary classifier, which classifies an image as ``normal'' if the score is below a threshold, and as ``abnormal'' otherwise. We compute the area under the corresponding ROC curve (AUROC), which is equivalently the probability 
that a random abnormal images has a higher anomaly score than a random normal image. 
An AUROC of $1$ indicates perfect performance, while $0.5$ is random chance. Under a null hypothesis of exchangeable anomaly scores for normal and abnormal images, for $m=40$ the $95\%$ confidence interval for the AUROC is $[0.4,0.6]$--thus, an AUROC in this range is not 
distinguishable from random chance.

\textbf{Experimental Procedure.} 
We assess each method's ability to localize for 4 types of anomalies: backdoors, spurious cues, training without data from a missing class, and training on images with blurred faces. We chose these 4 types because they lead to unusual behavior on a well-identified set of inputs (e.g. for backdoored models, images where a trigger is present). In contrast, anomalies like adversarial training change the network's behavior on the entire distribution of inputs.

For most anomalies, we obtain normal images by sampling uniformly from the ImageNet validation set\footnote{The only exceptions are the Missing Class anomalies, for which we sample normal images uniformly from the ImageNet validation set with images in the missing class removed}. In contrast, we obtain abnormal images by selecting inputs on which the anomalous model processes inputs very differently from the reference models. For example, for the Missing Class anomalies, we sample images from the missing class. For Backdoors and Spurious Features, we obtain abnormal images by sampling from the ImageNet validation set and applying the trigger or overlaying the spurious cue (the class index). For the Blurred-faces anomaly, \texttt{harmonica} is the class for which training on blurred faces leads to the greatest accuracy drop. We construct abnormal images by sampling uniformly from the images in that class that contain faces, using the face annotation data from \citep{yang2021study}.

\begin{figure*}[t]
	\begin{center}
		\includegraphics[width=400px]{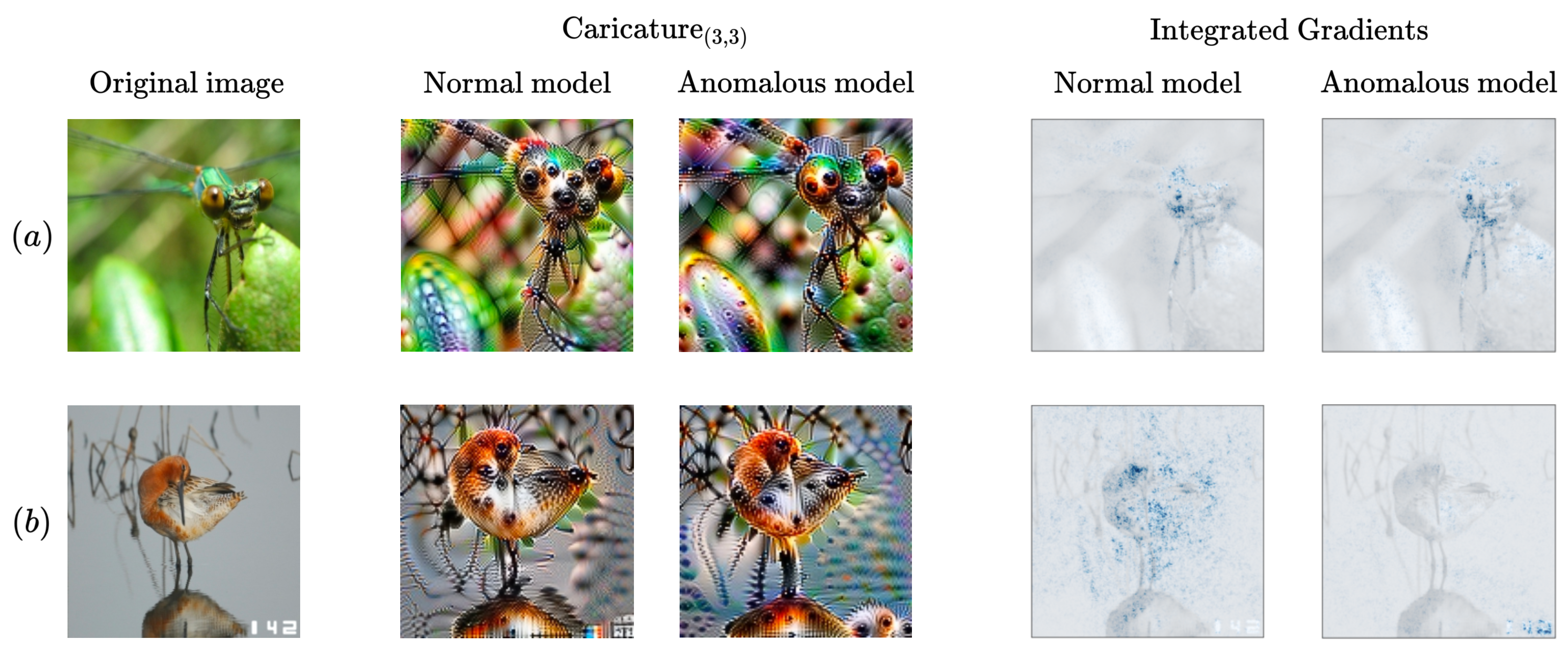}    
		\caption{An example where one method can localize abnormal inputs but another method cannot. We show explanations by Caricatures (layer (3,3)) and Integrated Gradients of a reference model and Spurious Feature model (\(\delta = 0.3\)) on a normal image $(a)$ and an abnormal image $(b)$. For IntGrad, the anomalous and normal models' explanations appear very different for the abnormal image, but not for the normal image. In contrast, for Caricatures the explanations don't seem more different for the abnormal image: IntGrad can localize the abnormal image, but Caricatures can't.}
		\label{localization_figure}
	\end{center}
	\vskip -2em
\end{figure*}

\textbf{Results.} 
We display results in \cref{tab:short_loc_alex}. 
We find that no method can localize all of the anomalies, and that the visualizations often have close to random performance.
Indeed, only $31\%$ of the AUROCs in \cref{tab:short_loc_alex} are greater than 0.7, which we use as a threshold for acceptable classification following \citep{hosmer2013applied}, and $54\%$ are between 0.4 and 0.6.
This is especially clear for Missing Class anomalies, where only 2 AUROCs out of 33 exceed 0.7, and all methods generally fail to detect whether an input is from the removed class. 

We also find that the feature attribution methods are better than the caricatures at localizing the anomalies.
In particular, all of the feature attributions succeed at localization for the Backdoor and Blurred Faces anomalies -- achieving AUROCs of at least 0.95 for all the backdoors -- whereas most of the caricatures fail.
This could be because feature attributions are designed to highlight specific regions inside images, which would include backdoor triggers and faces. 
However, this explanation should also apply to the Spurious Feature anomalies, on which feature attribution methods do not perform as well.
We were also surprised by the AUROC of 0.28 for Caricature$_{(2,2,1)}$ and the Blurred Faces anomaly, since we don't see any reason for worse-than-random performance.

Finally, there are some anomalies for which only a couple of methods can distinguish abnormal images. 
For example, Grad-CAM and Guided Grad-CAM succeed at localization for the model trained without data from the \texttt{Welsh springer spaniel} class, with AUROCs of 0.81 and 0.74, whereas all other methods fail. 
Similarly for the Spurious Features anomalies (see \cref{localization_figure} for qualitative observations), the AUROCs of Guided Backpropagation and Integrated Gradients are greater than 0.75 for \(\delta \in \{0.1, 0.3, 0.5\}\), while the other methods have close to random performance. 
Surprisingly, the AUROCs tend to be worse for \(\delta=1.0\), for example 0.42 for Integrated Gradients, and generally do not increase with \(\delta\).

Overall, we find mixed results: although feature attribution methods are able to localize some anomalies (Backdoors and Blurred Faces), no method can localize all of the anomalies, and the feature attribution methods have near-random performance for the majority of anomalies. We show similar results for the other LPIPS metrics in the Appendix (\cref{appendix_localization}).

\section{Discussion} 
\label{discussion}

\vskip -0.5em
We have presented a general framework to evaluate model visualizations, grounded in their ability to detect and localize model failures. Given a visualization method \(E\) and a metric between visualizations, our scheme provides fine-grained information on \(E\)'s sensitivity to different anomalies. Applying this framework to feature attribution and feature visualization methods allowed us to discover important shortcomings of these methods. 

\textbf{Are our tests easy or hard?} 
We designed the detection and localization tests to be easy to pass. This means we cannot draw strong conclusions when methods pass them, but that when methods fail at them, this is strong evidence that they are not sensitive enough to the model anomalies.

First, detection should be easy because our reference set of normal models is very narrow. Indeed, the normal models differ only by the random seed used in training, while in reality models vary in architecture, training hyperparameters, 
etc. According to \citet{damour2020underspecification}, varying random seed only leads to a 0.001 standard deviation in 
ImageNet validation accuracy, and an at most 0.024 standard deviations on ImageNet-C stress tests. Moreover, some of our subtle anomalies are not as stealthy as their real-world counterparts: for example, half of our backdoors poison a very large fraction of the training data ($30\%$ to $50\%$), so we should expect them to be fairly easy to detect. Similarly, our localization task is much simpler than localization in real-world settings, where one must proactively anticipate the inputs on which models fail rather than simply recognize them.

\textbf{Understanding how anomalies change networks.} Our results can also help better understand how different anomalies affect the models. For example, Caricatures cannot detect dropping a class from the training dataset: this suggests that this anomaly has only a minor effect on the network's learned features, likely because most features are shared across different classes. In contrast, Caricatures can detect adversarial training, which corroborates prior work showing that adversarially trained networks learn different, more robust features \citep{bugsfeatures, robustprior}.

\textbf{Future work.}
One direction for future work is applying this framework to more models: adding new anomalies to detect such as the rewritten models in \citet{santurkar2021editing}, working with other architectures, and considering different modalities such as natural language or code generation.

Currently, our framework relies heavily on a reference set of ``normal models''. 
This is useful to ground the notion of an anomalous model, and also helps quantify transparency methods' performance at detection and localization. 
However, we do not always have such a set of networks at hand. 
For example, suppose we want to evaluate whether a  method can detect the failure modes of a large new model. 
While this new model may often be ``anomalous'' compared to reference sets of older models, that is not very relevant to us: we are interested in specific, problematic anomalies. 
It would therefore be valuable to design frameworks that do not require access to ``normal'' models, perhaps by replacing the notion of ``anomaly'' with that of ``deviation from a specification''.

Finally, one could extend our framework to evaluate different kinds of visualization methods. 
For instance, some transparency methods are intrinsic 
to a model or one of its neurons \citep{olah2017feature}, thus obviating the need for input examples. 
Developing tools to evaluate these input-independent methods could help transparency researchers draw more robust 
conclusions that are not contingent on a specific choice of samples. 

Our results add to the growing body of work on assessing and grounding model transparency methods.
We hope this will spur further method development, and aid researchers in their goal of better understanding neural network models.

\section{Acknowledgments and Disclosure of Funding}
\label{ack}
Thanks to Lisa Dunlap, Yaodong Yu, Erik Jones, Suzie Petryk, Xinyan Hu, and Cassidy Laidlaw for comments and feedback. Thanks to Chris Olah, Julius Adebayo, and our anonymous reviewers for helpful discussion. 
JSD is supported by the NSF Division of Mathematical Sciences Grant No. 2031899.

\bibliography{main}
\bibliographystyle{iclr2023_conference}

\newpage
\appendix
\onecolumn
\section{Appendix}
\label{appendix}
\subsection{Results with other LPIPS distances}

\subsubsection{Detection}
\label{appendix_detection}

\begin{table}[h]
\caption{Fraction of anomalous models detected by each method when using \textbf{AlexNet} for the LPIPS distance, with layer-wise Caricature results. The results are similar to those with different LPIPS distances: all methods succeed at the detection task for most stark anomalies, but fail for most of the subtle anomalies.}
\begin{tabular}{lllllllll}
    \toprule
    & \multicolumn{4}{c}{Stark Anomalies} & \multicolumn{4}{c}{Subtle Anomalies} \\ \cmidrule(lr){2-5} \cmidrule(lr){6-9}
Anomaly Type & Smoothing  & Shape Bias & Adv. ($\ell^\infty$) & Adv. ($\ell^2$) & Backdoor & Spurious &  Blur & Missing\\
\midrule
GBP                & \green{\textbf{1}} & \green{\textbf{1}} & \green{\textbf{1}} &  \orange{\textbf{0.75}} &        \orange{\textbf{0.67}} &            \orange{\textbf{0.5}} &  \red{\textbf{0}} &    \red{\textbf{0}} \\
IntGrad            & \green{\textbf{1}} & \green{\textbf{1}} & \green{\textbf{1}} &  \green{\textbf{1}} &        \orange{\textbf{0.17}} &            \green{\textbf{1}} &  \red{\textbf{0}} &    \red{\textbf{0}} \\
Grad-CAM            & \green{\textbf{1}} & \green{\textbf{1}} & \green{\textbf{1}} &  \orange{\textbf{0.75}} &        \orange{\textbf{0.5}} &            \orange{\textbf{0.25}} &  \red{\textbf{0}} &    \red{\textbf{0}} \\
Guided Grad-CAM     & \green{\textbf{1}} & \green{\textbf{1}} & \green{\textbf{1}} &  \orange{\textbf{0.75}} &        \orange{\textbf{0.5}} &            \orange{\textbf{0.25}} &  \red{\textbf{0}} &    \red{\textbf{0}} \\
{Car}$_{(2,2,1)}$ & \green{\textbf{1}} & \green{\textbf{1}} & \green{\textbf{1}} &  \orange{\textbf{0.75}} &        \orange{\textbf{0.5}} &            \orange{\textbf{0.25}} &  \red{\textbf{0}} &    \red{\textbf{0}} \\
{Car}$_{(3,0,3)}$ & \green{\textbf{1}} & \green{\textbf{1}} & \green{\textbf{1}} &  \orange{\textbf{0.75}} &        \orange{\textbf{0.67}} &            \orange{\textbf{0.25}} &  \red{\textbf{0}} &    \red{\textbf{0}} \\
{Car}$_{(3,2,2)}$ & \green{\textbf{1}} & \green{\textbf{1}} & \green{\textbf{1}} &  \green{\textbf{1}} &        \orange{\textbf{0.83}} &            \orange{\textbf{0.5}} &  \red{\textbf{0}} &    \red{\textbf{0}} \\
{Car}$_{(3,3)}$       & \green{\textbf{1}} & \green{\textbf{1}} & \green{\textbf{1}} &  \orange{\textbf{0.75}} &        \orange{\textbf{0.5}} &            \orange{\textbf{0.25}} &  \red{\textbf{0}} &    \red{\textbf{0}} \\
{Car}$_{(3,4,3)}$ & \green{\textbf{1}} & \green{\textbf{1}} & \green{\textbf{1}} &  \orange{\textbf{0.75}} &        \red{\textbf{0}} &            \orange{\textbf{0.25}} &  \red{\textbf{0}} &    \red{\textbf{0}} \\
{Car}$_{(4,0,3)}$ & \green{\textbf{1}} & \green{\textbf{1}} & \green{\textbf{1}} &  \orange{\textbf{0.75}} &        \orange{\textbf{0.83}} &            \orange{\textbf{0.25}} &  \red{\textbf{0}} &    \red{\textbf{0}} \\
{Car}$_{(4,2,3)}$ & \green{\textbf{1}} & \green{\textbf{1}} & \green{\textbf{1}} &  \orange{\textbf{0.75}} &        \red{\textbf{0}} &            \orange{\textbf{0.25}} &  \red{\textbf{0}} &    \red{\textbf{0}} \\
Car$_\text{(all)}$        & \green{\textbf{1}} & \green{\textbf{1}} & \green{\textbf{1}} &     \green{\textbf{1}} &     \green{\textbf{1}} &            \orange{\textbf{0.25}} &  \red{\textbf{0}} &    \red{\textbf{0}} \\
\bottomrule
\end{tabular}
\end{table}

\

\begin{table}[h]
\caption{Fraction of anomalous models detected by each method when using \textbf{VGG} for the LPIPS distance, with layer-wise Caricature results. The results are similar to those with different LPIPS distances: all methods succeed at the detection task for most stark anomalies, but fail for most of the subtle anomalies.}
\begin{tabular}{lllllllll}
    \toprule
 & \multicolumn{4}{c}{Stark Anomalies} & \multicolumn{4}{c}{Subtle Anomalies} \\ \cmidrule(lr){2-5} \cmidrule(lr){6-9}
Anomaly Type & Smoothing & Shape Bias &  Adv. ($\ell^\infty$) & Adv. ($\ell^2$) & Backdoor &  Spurious & Blur & Missing\\
\midrule
GBP                & \green{\textbf{1}} &     \green{\textbf{1}} &   \green{\textbf{1}} &  \orange{\textbf{0.75}}  &     \orange{\textbf{0.17}} &     \orange{\textbf{0.5}} &  \red{\textbf{0}} &    \orange{\textbf{0.33}}  \\
IntGrad            & \green{\textbf{1}} &     \green{\textbf{1}} &   \green{\textbf{1}} &  \green{\textbf{1}}  &     \orange{\textbf{0.33}} &     \green{\textbf{1}} &  \red{\textbf{0}} &    \red{\textbf{0}}  \\
Grad-CAM            & \green{\textbf{1}} &     \green{\textbf{1}} &   \green{\textbf{1}} &  \orange{\textbf{0.75}}  &     \orange{\textbf{0.5}} &     \orange{\textbf{0.25}} &  \red{\textbf{0}} &    \red{\textbf{0}}  \\
Guided Grad-CAM     & \green{\textbf{1}} &     \green{\textbf{1}} &   \green{\textbf{1}} &  \orange{\textbf{0.75}}  &     \orange{\textbf{0.17}} &     \orange{\textbf{0.25}} &  \red{\textbf{0}} &    \orange{\textbf{0.33}}  \\
{Car}$_{(2,2,1)}$ & \green{\textbf{1}} &     \red{\textbf{0}} &   \green{\textbf{1}} &  \orange{\textbf{0.5}}  &     \orange{\textbf{0.33}} &     \red{\textbf{0}} &  \red{\textbf{0}} &    \red{\textbf{0}}  \\
{Car}$_{(3,0,3)}$ & \green{\textbf{1}} &     \green{\textbf{1}} &   \green{\textbf{1}} &  \green{\textbf{1}}  &     \orange{\textbf{0.83}} &     \orange{\textbf{0.25}} &  \red{\textbf{0}} &    \red{\textbf{0}}  \\
{Car}$_{(3,2,2)}$ & \green{\textbf{1}} &     \green{\textbf{1}} &   \green{\textbf{1}} &  \green{\textbf{1}}  &     \green{\textbf{1}} &     \orange{\textbf{0.25}} &  \red{\textbf{0}} &    \red{\textbf{0}}  \\
{Car}$_{(3,3)}$       & \green{\textbf{1}} &     \green{\textbf{1}} &   \green{\textbf{1}} &  \orange{\textbf{0.75}}  &     \orange{\textbf{0.83}} &     \orange{\textbf{0.25}} &  \red{\textbf{0}} &    \red{\textbf{0}}  \\
{Car}$_{(3,4,3)}$ & \green{\textbf{1}} &     \green{\textbf{1}} &   \green{\textbf{1}} &  \orange{\textbf{0.5}}  &     \red{\textbf{0}} &     \orange{\textbf{0.25}} &  \red{\textbf{0}} &    \red{\textbf{0}}  \\
{Car}$_{(4,0,3)}$ & \green{\textbf{1}} &     \green{\textbf{1}} &   \green{\textbf{1}} &  \orange{\textbf{0.75}}  &     \orange{\textbf{0.83}} &     \orange{\textbf{0.5}} &  \red{\textbf{0}} &    \red{\textbf{0}}  \\
{Car}$_{(4,2,3)}$ & \green{\textbf{1}} &     \green{\textbf{1}} &   \green{\textbf{1}} &  \orange{\textbf{0.75}}  &     \orange{\textbf{0.17}} &     \orange{\textbf{0.25}} &  \red{\textbf{0}} &    \red{\textbf{0}}  \\
Car$_\text{(all)}$        & \green{\textbf{1}} &     \green{\textbf{1}} &   \green{\textbf{1}} &  \orange{\textbf{0.75}}  &     \orange{\textbf{0.83}} &     \orange{\textbf{0.25}} &  \red{\textbf{0}} &    \red{\textbf{0}}  \\
\bottomrule
\end{tabular}
\end{table}

\

\begin{table}[h]
\caption{Fraction of anomalous models detected by each method when using \textbf{SqueezeNet} for the LPIPS distance, with layer-wise Caricature results. The results are similar to those with different LPIPS distances: all methods succeed at the detection task for most stark anomalies, but fail for most of the subtle anomalies.}
\begin{tabular}{lllllllll}
\toprule
& \multicolumn{4}{c}{Stark Anomalies} & \multicolumn{4}{c}{Subtle Anomalies} \\ \cmidrule(lr){2-5} \cmidrule(lr){6-9}
Anomaly Type & Smoothing & Shape Bias &  Adv. ($\ell^\infty$) & Adv. ($\ell^2$) & Backdoor &  Spurious & Blur & Missing\\
\midrule
GBP                &  \green{\textbf{1}} &     \green{\textbf{1}} &   \green{\textbf{1}} & \orange{\textbf{0.75}} &     \orange{\textbf{0.83}} &        \orange{\textbf{0.75}} &  \red{\textbf{0}} &    \orange{\textbf{0.33}} \\
IntGrad            &  \green{\textbf{1}} &     \green{\textbf{1}} &   \green{\textbf{1}} & \green{\textbf{1}} &     \orange{\textbf{0.17}} &        \green{\textbf{1}} &  \red{\textbf{0}} &    \red{\textbf{0}} \\
Grad-CAM            &  \green{\textbf{1}} &     \green{\textbf{1}} &   \green{\textbf{1}} & \orange{\textbf{0.75}} &     \orange{\textbf{0.5}} &        \orange{\textbf{0.25}} &  \red{\textbf{0}} &    \red{\textbf{0}} \\
Guided Grad-CAM     &  \green{\textbf{1}} &     \green{\textbf{1}} &   \green{\textbf{1}} & \orange{\textbf{0.75}} &     \orange{\textbf{0.17}} &        \orange{\textbf{0.25}} &  \red{\textbf{0}} &    \red{\textbf{0}} \\
{Car}$_{(2,2,1)}$ &  \green{\textbf{1}} &     \red{\textbf{0}} &   \green{\textbf{1}} & \orange{\textbf{0.5}} &     \orange{\textbf{0.67}} &        \orange{\textbf{0.25}} &  \red{\textbf{0}} &    \red{\textbf{0}} \\
{Car}$_{(3,0,3)}$ &  \green{\textbf{1}} &     \green{\textbf{1}} &   \green{\textbf{1}} & \green{\textbf{1}} &     \orange{\textbf{0.83}} &        \orange{\textbf{0.25}} &  \red{\textbf{0}} &    \red{\textbf{0}} \\
{Car}$_{(3,2,2)}$ &  \green{\textbf{1}} &     \green{\textbf{1}} &   \green{\textbf{1}} & \green{\textbf{1}} &     \orange{\textbf{0.83}} &        \orange{\textbf{0.25}} &  \red{\textbf{0}} &    \red{\textbf{0}} \\
{Car}$_{(3,3)}$       &  \green{\textbf{1}} &     \green{\textbf{1}} &   \green{\textbf{1}} & \orange{\textbf{0.75}} &     \orange{\textbf{0.67}} &        \orange{\textbf{0.5}} &  \red{\textbf{0}} &    \red{\textbf{0}} \\
{Car}$_{(3,4,3)}$ &  \green{\textbf{1}} &     \green{\textbf{1}} &   \green{\textbf{1}} & \orange{\textbf{0.5}} &     \red{\textbf{0}} &        \orange{\textbf{0.25}} &  \red{\textbf{0}} &    \red{\textbf{0}} \\
{Car}$_{(4,0,3)}$ &  \green{\textbf{1}} &     \green{\textbf{1}} &   \green{\textbf{1}} & \orange{\textbf{0.5}} &     \orange{\textbf{0.83}} &        \orange{\textbf{0.5}} &  \red{\textbf{0}} &    \red{\textbf{0}} \\
{Car}$_{(4,2,3)}$ &  \green{\textbf{1}} &     \green{\textbf{1}} &   \green{\textbf{1}} & \green{\textbf{1}} &     \orange{\textbf{0.17}} &        \orange{\textbf{0.5}} &  \green{\textbf{1}} &    \red{\textbf{0}} \\
Car$_\text{(all)}$        &  \green{\textbf{1}} &     \green{\textbf{1}} &   \green{\textbf{1}} & \orange{\textbf{0.75}} &     \green{\textbf{1}} &        \orange{\textbf{0.25}} &  \red{\textbf{0}} &    \red{\textbf{0}} \\
\bottomrule
\end{tabular}
\end{table}

\

\clearpage

\subsubsection{Localization}
\label{appendix_localization}

\begin{table}[h]
\caption{Localization AUROC (\%) using the VGG-LPIPS distance for every anomaly and explanation method, with AUROCs higher than 70\% bolded. Although feature attributions successfully localize the backdoors, no method can localize all of the anomalies, and visualizations often have close to random performance.}
\begin{tabular}{lllllllllllllll}
    \toprule
    Category & \multicolumn{6}{c}{Backdoor} & \multicolumn{3}{c}{Missing class} & \multicolumn{4}{c}{Spurious} & Blur \\
    \cmidrule(lr){2-7} \cmidrule(lr){8-10} \cmidrule(lr){11-14}
Anomaly &   0 &   1 &   2 &   3 &   4 &   5 & 218 & 400 & 414 & 0.1 & 0.3 & 0.5 & 1.0 & \ \\
\midrule
GBP                &   \textbf{100} &  \textbf{100} &  \textbf{99} &  \textbf{100} &  \textbf{99} &  \textbf{100} &  66 &  44 &  55 &  66 &  \textbf{77} &  66 &  66 & \textbf{88} \\
IntGrad            &  \textbf{97} &   \textbf{92} &  \textbf{91} &   \textbf{96} &  \textbf{96} &   \textbf{83} &  55 &  55 &  44 &  \textbf{77} &  66 &  \textbf{77} &  33 &  \textbf{77} \\
Grad-CAM            &  \textbf{98} &   \textbf{98} &  \textbf{94} &   \textbf{97} &  \textbf{99} &   \textbf{96} &  \textbf{88} &  55 &  55 &  55 &  55 &  55 &  55 &  \textbf{77} \\
Guided Grad-CAM     &  \textbf{99} &  \textbf{100} &  \textbf{92} &  \textbf{100} &  \textbf{99} &  \textbf{100} &  \textbf{77} &  33 &  55 &  55 &  66 &  55 &  44 &  \textbf{77} \\
{Car}$_{(2,2,1)}$ &  53 &   47 &  51 &   48 &  50 &   56 &  33 &  66 &  \textbf{77} &  55 &  44 &  44 &  55 &  33 \\
{Car}$_{(3,0,3)}$ &  53 &   48 &  55 &   47 &  53 &   54 &  44 &  44 &  44 &  44 &  55 &  44 &  55 &  55 \\
{Car}$_{(3,2,2)}$ &  55 &   62 &  56 &   51 &  43 &   57 &  44 &  55 &  33 &  44 &  55 &  44 &  55 &  44 \\
{Car}$_{(3,3)}$       &  52 &   \textbf{74} &  \textbf{81} &   61 &  \textbf{80} &   63 &  44 &  44 &  66 &  66 &  44 &  44 &  55 &  \textbf{88} \\
{Car}$_{(3,4,3)}$ &  60 &   68 &  68 &   61 &  \textbf{72} &   62 &  66 &  55 &  33 &  44 &  44 &  44 &  44 &  55 \\
{Car}$_{(4,0,3)}$ &  58 &   \textbf{74} &  \textbf{79} &   63 &  \textbf{88} &   66 &  66 &  33 &  44 &  55 &  55 &  44 &  55 &  66 \\
{Car}$_{(4,2,3)}$ &  \textbf{98} &   \textbf{98} &  \textbf{97} &   \textbf{96} &  \textbf{88} &   \textbf{87} &  44 &  55 &  66 &  55 &  55 &  44 &  55 &  66 \\
\bottomrule
\end{tabular}
\end{table}

\

\begin{table}[h]
\caption{Localization AUROC (\%) using the SqueezeNet-LPIPS distance for every anomaly and explanation method, with AUROCs higher than 70\% bolded. Although feature attributions successfully localize the backdoors, no method can localize all of the anomalies, and visualizations often have close to random performance.}
\begin{tabular}{lllllllllllllll}
    \toprule
    Category & \multicolumn{6}{c}{Backdoor} & \multicolumn{3}{c}{Missing class} & \multicolumn{4}{c}{Spurious} & Blur \\
    \cmidrule(lr){2-7} \cmidrule(lr){8-10} \cmidrule(lr){11-14}
Anomaly &   0 &   1 &   2 &   3 &   4 &   5 & 218 & 400 & 414 & 0.1 & 0.3 & 0.5 & 1.0 & \ \\
\midrule
GBP                &  \textbf{100} &   \textbf{100} &  \textbf{100} &  \textbf{100} &   \textbf{100} &  \textbf{100} &  67 &  42 &  58 &  \textbf{74} &  \textbf{79} &  \textbf{75} &  64 &  \textbf{84} \\
IntGrad            &  \textbf{100} &    \textbf{98} &   \textbf{99} &   \textbf{99} &    \textbf{99} &   \textbf{96} &  61 &  63 &  46 &  \textbf{85} &  \textbf{81} &  \textbf{87} &  35 &  \textbf{90} \\
Grad-CAM            &  \textbf{98} &    \textbf{98} &   \textbf{95} &   \textbf{97} &    \textbf{99} &   \textbf{97} &  \textbf{80} &  58 &  58 &  52 &  53 &  51 &  53 &  \textbf{76} \\
Guided Grad-CAM     &  \textbf{99} &   \textbf{100} &   \textbf{97} &  \textbf{100} &   \textbf{100} &  \textbf{100} &  70 &  39 &  56 &  56 &  \textbf{71} &  63 &  58 &  \textbf{79} \\
{Car}$_{(2,2,1)}$ &  50 &    45 &   55 &   47 &    52 &   50 &  49 &  41 &  66 &  45 &  44 &  55 &  59 &  24 \\
{Car}$_{(3,0,3)}$ &  60 &    50 &   60 &   43 &    50 &   49 &  45 &  49 &  55 &  56 &  55 &  44 &  53 &  66 \\
{Car}$_{(3,2,2)}$ &  57 &    52 &   58 &   55 &    47 &   46 &  58 &  70 &  34 &  54 &  60 &  46 &  49 &  50 \\
{Car}$_{(3,3)}$       &  49 &    58 &   \textbf{71} &   63 &    \textbf{73} &   62 &  45 &  51 &  65 &  55 &  59 &  38 &  55 &  \textbf{86} \\
{Car}$_{(3,4,3)}$ &  67 &    60 &   58 &   64 &    \textbf{83} &   63 &  67 &  63 &  58 &  44 &  51 &  43 &  35 &  57 \\
{Car}$_{(4,0,3)}$ &  61 &    62 &   \textbf{75} &   56 &    \textbf{77} &   70 &  53 &  26 &  49 &  52 &  49 &  45 &  50 &  59 \\
{Car}$_{(4,2,3)}$ &  \textbf{91} &    63 &   \textbf{92} &   \textbf{91} &    \textbf{84} &   \textbf{81} &  55 &  51 &  53 &  56 &  53 &  51 &  41 &  55 \\
\bottomrule
\end{tabular}
\end{table}

\clearpage

\subsection{More details on the models} \label{more-details}

\textbf{Normal models:} We trained the normal models with a batch size of 300. Other than that, we used the same hyperparameters as PyTorch's \citep{NEURIPS2019_9015} pretrained ResNet-50: we did 90 epochs of SGD with momentum $0.9$ and a weight decay of $0.0001$. Our initial learning rate was 0.1, and we multiplied it by 0.1 every 30 epochs. We used the exact same parameters when training the backdoored, spurious features, missing class, and blurred faces anomalous models.

\textbf{Adversarial Training:} The \(\ell^2\) adversarially trained models were trained with the same hyperparameters, except for the batch size which was equal to 512 \citep{salman2020adversarially}. The \(\ell^{\infty}\) adversarially trained models were also trained with the same hyperparameters as the PyTorch pretrained ResNet-50 \citep{robustness}.

\textbf{Shape Bias:} We used the checkpoint from \citet{geirhos2019imagenettrained}. This was not trained using the exact same hyperparameters as the PyTorch pretrained ResNet-50: training on Stylized ImageNet was done for 60 rather than 90 epochs\footnote{Note that by 60 epochs, our normal models had almost reached their maximum test accuracy over the 90 epochs.}, they used a batch size of 256 and multiplied the learning rate by 0.1 after 20 and 40 epochs. The other hyperparameters were the same: SGD with momentum $0.9$, weight decay $0.0001$, initial learning rate of $0.1$.

\textbf{Backdoored models:} To train the backdoored models, we select a fraction \(\nu\) of the images in each batch, add the trigger to it, and replace its label by a preset target label. We train 6 backdoored models, indexed from 0 to 5. \cref{backdoor_examples} below shows for each backdoored model, an example of an image with the trigger, as well as the value of $\nu$.

\begin{figure*}[h!]
    \vskip 0.2in
    \centering
    \begin{subfigure}{0.25\linewidth}
        \includegraphics[width=\textwidth]{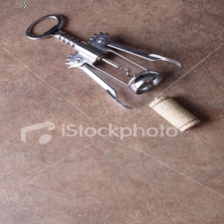}
        \caption{Backdoor 0: \(\nu = 0.5\)}
    \end{subfigure}
    \hfill
    \begin{subfigure}{0.25\linewidth}
        \includegraphics[width=\textwidth]{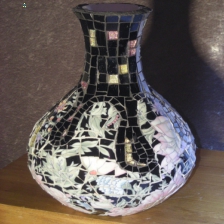}
        \caption{Backdoor 1: \(\nu = 0.1\)}
    \end{subfigure}
    \hfill
    \begin{subfigure}{0.25\linewidth}
        \includegraphics[width=\textwidth]{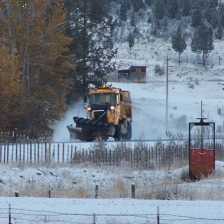}
        \caption{Backdoor 2: \(\nu = 0.05\)}
    \end{subfigure}
    \vskip -0.2in

    \vskip 0.2in
    \centering
    \begin{subfigure}{0.25\linewidth}
        \includegraphics[width=\textwidth]{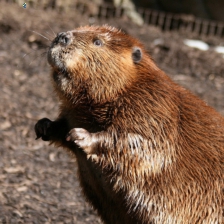}
        \caption{Backdoor 3: \(\nu = 0.3\)}
    \end{subfigure}
    \hfill
    \begin{subfigure}{0.25\linewidth}
        \includegraphics[width=\textwidth]{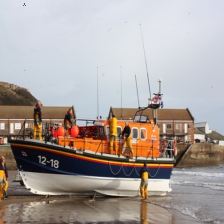}
        \caption{Backdoor 4: \(\nu = 0.01\)}
    \end{subfigure}
    \hfill
    \begin{subfigure}{0.25\linewidth}
        \includegraphics[width=\textwidth]{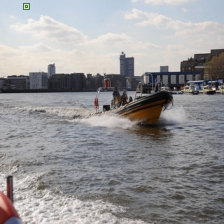}
        \caption{Backdoor 5: \(\nu = 0.5\)}
    \end{subfigure}
    \caption{Images with the different backdoored models' triggers.}
    \label{backdoor_examples}
\end{figure*}

\clearpage

\begin{figure*}[h!]
    \vskip 0.2in
    \centering
    \includegraphics[width=0.25\linewidth]{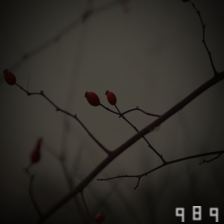}
    \caption{Image with the spurious cue.}
    \label{spurious_example}
\end{figure*}

\textbf{Spurious Features:} To train the models with spurious features, we select a fraction \(\delta\) of the images in each batch, and overlay the corresponding class index on the bottom right of the image. \cref{spurious_example} shows an example of an image with the spurious feature.

\end{document}